%% file: main.tex
\documentclass[10pt,twocolumn,letterpaper]{article}

\PassOptionsToPackage{dvipsnames,table}{xcolor} 

\usepackage{algorithm}
\usepackage{algpseudocode}
\usepackage{pgfplots}
\pgfplotsset{compat=1.18}
\usepackage{listings}
\usepackage{colortbl}
\usepackage{multirow}

\usepackage[numbers,sort&compress]{natbib}
\usepackage[pagenumbers]{iccv} 


\definecolor{iccvblue}{rgb}{0.21,0.49,0.74}
\usepackage[pagebackref,breaklinks,colorlinks,allcolors=iccvblue]{hyperref}

\definecolor{navyblue}{RGB}{191, 209, 229} 
\definecolor{light_yellow}{RGB}{255,243,194}
\definecolor{orange}{RGB}{255,200,100}

\def\paperID{****} 
\def\confName{****}
\def\confYear{****}

\title{SG-Tailor: Inter-Object Commonsense Relationship Reasoning \\ for Scene Graph Manipulation}

\author{
Haoliang Shang$^{1,*}$ \hspace{1.2ex} Hanyu Wu$^{1,*}$ \hspace{1.2ex} Guangyao Zhai$^{1,2,3}$ \hspace{1.2ex} Boyang Sun$^1$ \hspace{1.2ex} Fangjinhua Wang$^1$\\[0.2ex]
Federico Tombari$^{2,4}$ \hspace{1.3ex} Marc Pollefeys$^{1,5}$\\[1.5ex]
$^1$ ETH Zurich \quad $^2$ TU Munich \quad $^3$ MCML \quad $^4$ Google \quad $^5$ Microsoft\\[0.2ex]
{\tt\small \{hshang,hanywu\}@ethz.ch}
\quad {\tt\small \{guangyao.zhai,federico.tombari\}@tum.de}\\
\quad {\tt\small \{boyang.sun,fangjinhua.wang,marc.pollefeys\}@inf.ethz.ch}
}

\begin{document}

\twocolumn[{
  \maketitle
  \begin{center}
    \includegraphics[width=0.95\linewidth]{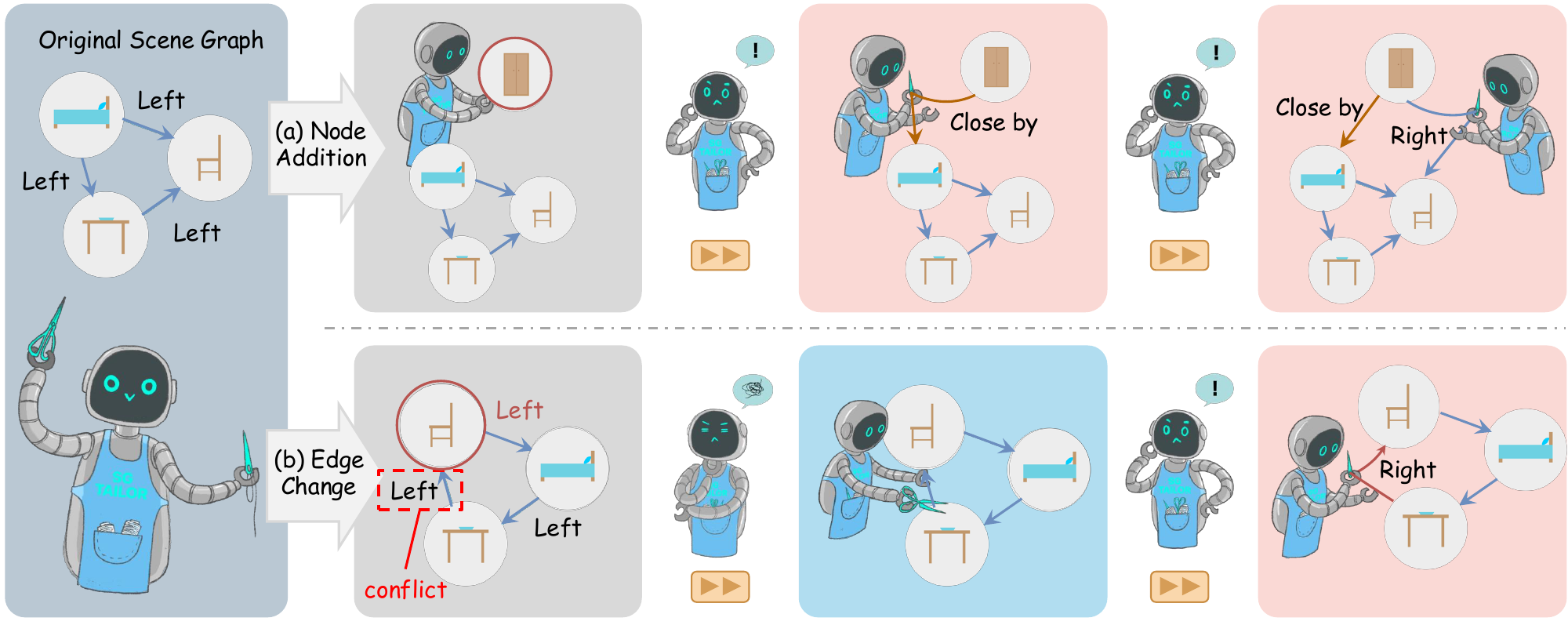}
    \vspace{0.5em}
    \captionof{figure}{
    \textbf{SG-Tailor for scene graph manipulation.} SG-Tailor manipulates a given scene graph in two modes: (a) \textit{Node Addition} and (b) \textit{Edge Change}. For node addition, SG-Tailor autoregressively \textbf{reasons commonsense relationships} between a newly introduced node and existing nodes (\textit{e.g.}, The wardrobe node should be near the bed node and to the left of the chair node, as the chair is already left of the bed.). For edge change, it maintains the desired edge while \textbf{resolving conflicts} (\textit{e.g.}, naively moving the chair node to the left of the bed node causes a table conflict. SG-Tailor resolves this by replacing the conflicting edge to maintain coherence).
    }
    \label{fig:teaser}
  \end{center}
}]

\input{sec/0_abstract}    
\input{sec/1_intro}
\input{sec/2_related_work}
\input{sec/3_preliminaries}

\input{sec/4_problem_formulation}

\input{sec/5_method}

\input{sec/6_experiment}

\input{sec/7_downstream_task}
\input{sec/8_conclusion}

\newpage
{
    \small
   \bibliographystyle{unsrtnat}
    \bibliography{main}
}

\appendix
\input{sec/supplementary}

\end{document}

%% file: sec/0_abstract.tex
\begin{abstract}
Scene graphs capture complex relationships among objects, serving as strong priors for content generation and manipulation. Yet, reasonably manipulating scene graphs -- whether by adding nodes or modifying edges -- remains a challenging and untouched task. Tasks such as adding a node to the graph or reasoning about a node's relationships with all others are computationally intractable, as even a single edge modification can trigger conflicts due to the intricate interdependencies within the graph. To address these challenges, we introduce SG-Tailor, an autoregressive model that predicts the conflict-free relationship between any two nodes. SG-Tailor not only infers inter-object relationships, including generating commonsense edges for newly added nodes but also resolves conflicts arising from edge modifications to produce coherent, manipulated graphs for downstream tasks. For node addition, the model queries the target node and other nodes from the graph to predict the appropriate relationships. For edge modification, SG-Tailor employs a Cut-And-Stitch strategy to solve the conflicts and globally adjust the graph. Extensive experiments demonstrate that SG-Tailor outperforms competing methods by a large margin and can be seamlessly integrated as a plug-in module for scene generation and robotic manipulation tasks. The code will be here \footnote{\url{https://github.com/josef5838/SG-Tailor}}.

\end{abstract}

%% file: sec/1_intro.tex
\section{Introduction}
\label{sec:intro}
Scene graphs effectively capture semantic relationships among objects by representing them as nodes and their interactions as edges~\cite{li2024scene,chang2021comprehensive}. This structured, interpretable representation is widely used in computer vision tasks, such as image captioning~\cite{krishna2017visual}, scene understanding~\cite{zhang2021holistic,gu2024conceptgraphs}, and robotics applications ~\cite{chang2023hydra,hughes2022hydra,werby2024hierarchical,maggio2024clio,jiang2024roboexp}.

Additionally, scene graphs uniquely offer a bidirectional capability: they decompose scenes into detailed representations of objects and relationships~\cite{yang2018graph,xu2017scene,tang2020unbiased,dhamo2020semantic}, and can compose coherent 2D/3D content based on generative models for advanced applications~\cite{dhamo2020semantic,zhai2024echoscene,lin2024instructscene,gao2024graphdreamer}. 
Building on this foundation, a highly flexible pipeline is envisioned for data creation and manipulation using scene graphs. 
\textbf{First}, a general extractor embeds visual data into scene graphs.
\textbf{Next}, a graph manipulator modifies these graphs according to user commands, producing diverse yet commonsense-compliant configurations.
\textbf{Finally}, a generative model synthesizes realistic data from the manipulated graphs for various downstream applications.
This streamlined approach significantly enhances flexibility and precision in interactive data generation conditioned on the large amount of 3D scene content \cite{sverse,wald2020learning,ramakrishnan2021hm3d}.

While extensive research has addressed the initial graph extraction~\cite{scenegraphfusion,rosinol2021kimera,johnson2015image,im2024egtr} and the final generative synthesis~\cite{johnson2018image,dhamo2020semantic,commonscenes,yang2025mmgdreamer}, scene graph manipulation itself has been largely overlooked. Even though some methods~\cite{dhamo2020semantic,commonscenes,dhamo2021graph} apply manipulation inside their pipeline, they do not consider potential conflicts when changing semantics. However, manipulating scene graphs—whether through node addition or edge modification—is inherently challenging due to their complexity. Even minor changes can cascade through the graph, disrupting overall coherence. For example, adding a node requires not only introducing the new node but also reasoning about its plausible relationships with existing nodes, as shown in~\autoref{fig:teaser} (a). Identifying reasonable and commonsense relationships is a complex problem due to the inherent uncertainty and combinatorial complexity of potential relationship combinations. Similarly, modifying a single edge can disturb intricate inter-node dependencies, potentially introducing logical conflicts and making the graph incoherent (see~\autoref{fig:teaser} (b) for a case). 
Identifying and resolving these conflicts to maintain graph consistency remains an unexplored area. Such difficulty steers up when the complexity and scale of the scene graphs increase.

In this work, we propose \emph{SG-Tailor}, an autoregressive model designed to tackle the intricacies of scene graph manipulation. SG-Tailor operates by predicting the relationship between any two nodes in the context of an existing scene graph. This capability allows the model to infer commonsense edges for newly added nodes while ensuring that edge modifications do not introduce inconsistencies.
Specifically, for node additions, SG-Tailor queries the new node alongside existing nodes to infer their relationships, as depicted in~\autoref{fig:teaser} (a). This ensures seamless integration of new nodes into the current scene graph structure. For edge modification, the model employs a novel \emph{Cut-And-Stitch} strategy. First, SG-Tailor isolates the subject node from the graph by cutting off all linked edges, then it infers and "stitches" all relationships conditioned on the rest of the graph, thereby removing all possible conflicts in the graph. Through learning, SG-Tailor not only learns the spatial constraints but also the inter-object commonsense relationship reasoning to restore global coherence. In such a way, we bypass the computational complexity of detecting and resolving relationship conflicts, particularly in densely connected graphs.
We validate SG-Tailor through extensive experiments across diverse benchmarks. Results demonstrate that SG-Tailor significantly outperforms existing methods and serves as a robust plug-in module for downstream tasks, including scene generation and robotic manipulation. By effectively bridging the gap between theoretical complexity and practical utility, SG-Tailor paves new avenues for future research in the manipulation and application of structured visual representations.

Our contributions are summarized as follows:
\begin{itemize}
\item We reveal the overlooked problems of scene graph manipulation, highlighting the importance of maintaining semantic coherence during node and edge modifications.
\item We propose \emph{SG-Tailor}, an autoregressive model for robust scene graph manipulation capable of commonsense-aware relationship reasoning and conflict solving.
\item We demonstrate that SG-Tailor not only significantly outperforms existing competitors on diverse benchmarks but also proves its practical effectiveness as a plug-in module for scene generation and robotic manipulation tasks.
\end{itemize}

%% file: sec/2_related_work.tex
\section{Related work}

\paragraph{Scene Graphs}
Scene graphs, as symbolic and semantic representations~\cite{johnson2015image,krishna2017visual,armeni20193d}, can be obtained from texts~\cite{zhao2023textpsg}, 2D images~\cite{xu2017scene,zellers2018neural,qi2019attentive,herzig2018mapping}, 3D geometry~\cite{koch2024sgrec3d,rosinol2021kimera} and 4D data~\cite{yang20234d} for spatial and temporal understanding.
Scene graphs can facilitate various tasks, including  retrieval~\cite{johnson2015image},  generation~\cite{dhamo2020semantic,johnson2018image}, and VQA~\cite{teney2017graph}. More embodied applications include robotic manipulation~\cite{zhai2024sg,jiang2024roboexp}, and mobile navigation~\cite{rana2023sayplan}. Most work focuses on how to embed information into the graphs and how to generate concrete content from scene graphs, ignoring the importance of graph manipulation. In this work, we expose this overlooked problem in current work~\cite{,dhamo2021graph} and propose SG-Tailor to solve it. The closest work to ours is SGNet~\cite{scenegraphnet}, which predicts objects given the contextual scene graph and the target locations. In contrast, our framework focuses on reasoning inter-object relationships based on textual information without explicit geometric cues.

\paragraph{Autoregressive Models}
Autoregressive models sequentially predict the next component based on previous inputs. 
In earlier years, It has shown the possibility of generating  RGB in a row-by-row, raster-scan manner~\cite{van2016conditional,chen2020generative} in image generation. VQGAN~\cite{esser2021taming} further performs autoregressive learning in the latent space of VQVAE~\cite{van2017neural}. Recently, autoregressive models have dominated natural language processing (NLP), serving as an important component of Multimodal large language models~(MLLMs)~\cite{llama,brown2020language,alayrac2022flamingo,guo2025deepseek,liu2023visual}. Beyond text regression, autoregressive models are extended to perform robotics manipulation tasks, serving as a core for vision-language-action models~\cite{team2024octo,kim2024openvla,black2024pi_0}. Since scene graphs are natural textual information, we use autoregressive models to treat scene graph manipulation as a next-token prediction task inspired by MLLMs.

%% file: sec/3_preliminaries.tex
\begin{figure*}[t!] 
    \centering          
    \includegraphics[width=0.98\textwidth]{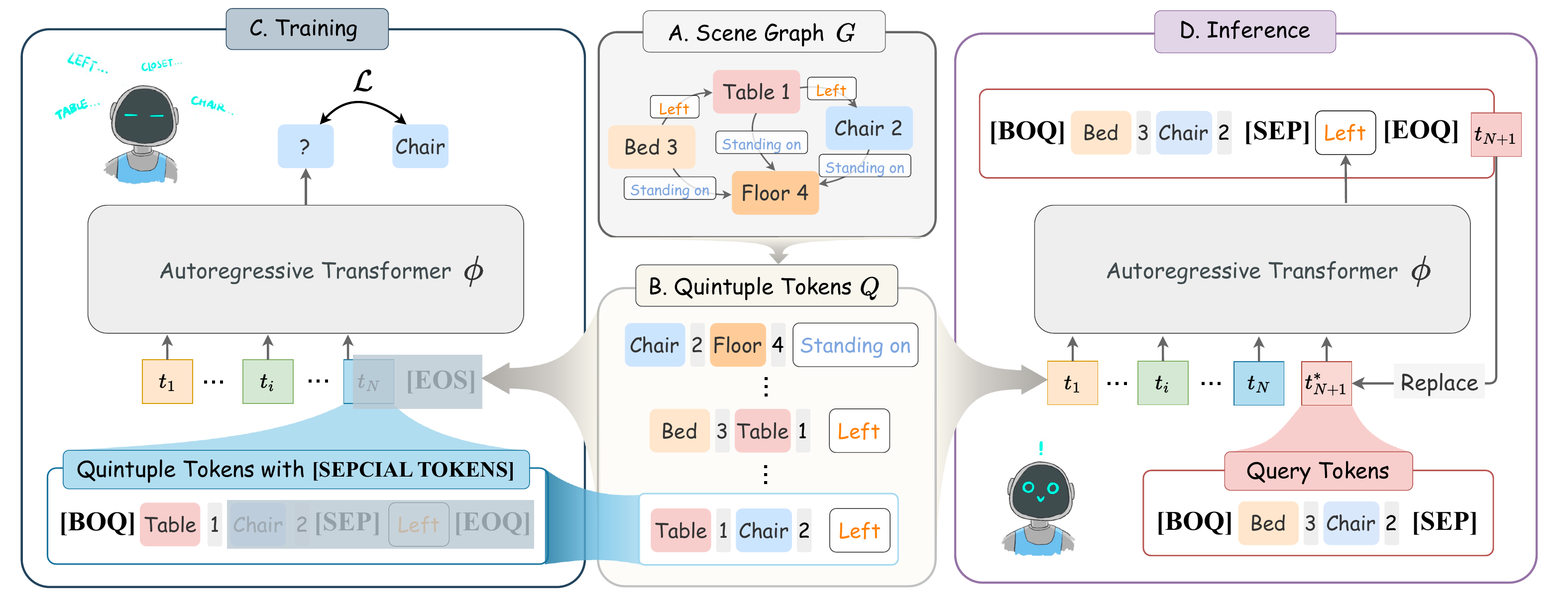} 
    \caption{\textbf{Training and Inference.} Starting from \emph{A. Scene Graph} $G$, we convert $N$ triplets into a set of \emph{B. Quintuple Tokens} $Q$, resulting in $5N$ tokens. Each token $q_i \in Q$ is then combined with special tokens to form token sequences $t_i$. During \emph{C. Training}, the model $\phi$ learns to perform next-token prediction, including both object nodes and relationships, with the mask attention mechanism. This process runs until it reaches the sequence-end token $\mathbf{[EOS]}$. During \emph{D. Inference}, $\phi$ accepts all existing tokens $\left\{t_1,...t_N\right\}$ from the given graph and query tokens $t_{N+1}^*$ containing each two of nodes and special tokens to perform next-relationship prediction. The predicted relationship is integrated into $t_{N+1}^*$, forming $t_{N+1}$. In this way, the model autoregressively reasons about inter-object relationships.
    }

    \label{fig:pipeline} 
\end{figure*}

%% file: sec/4_problem_formulation.tex
\section{Problem Formulation} \label{Problem}
We formulate the problem by first defining the set of \emph{Reasonable Scene Graphs} based on scene graph representations. 
Next, we categorize the operation classes of scene graph manipulation and conduct a comprehensive analysis of the associated challenges, demonstrating that these challenges can be decomposed into \emph{Cut} and \emph{Stitch} steps.
\subsection{Reasonable Scene Graph} \label{RSG}
The scene graph~\cite{li2024scene} used in this work is denoted by
\(G = \{V, E\},\)
which serves as a structured representation of a visual scene.
\(
V = \{ v_i \mid i = 1, \dots, M \},
\)
where each \(v_i\) is an object node, and
\(E = \{ e_{i\to j} \mid i, j = 1, \dots, M, \, i \neq j \},\)
where each \(e_{i\to j}\) is a directed edge from node \(v_i\) to node \(v_j\). Every node \(v_i\) and edge \(e_{i\to j}\) incorporate categorical information (class label) about the object and the instance index.
Based on \(G\), we formulate \emph{Reasonable Scene Graphs} as:
\begin{equation}
\mathbb{G} = \{\,G \in \mathcal{G}\mid G \models P_1 \wedge P_2 \,\},
\label{eq:reasonable_sg}
\end{equation}
where $\mathcal{G}$ denotes the set of all possible scene graphs, 
and $P_1, P_2$ denote commonsense constraints and spatial constraints respectively.
$\models$ indicates that the graph $G$ satisfies these constraints. All reasonable scene graphs have the optimal substructure property, i.e., for any $G \in \mathbb{G}$, every subgraph of $G$ is also reasonable.\label{property:optimal_substruct} 

\subsection{Scene Graph Manipulation} \label{sec:sgmod}
We formally define scene graph manipulation—the process of modifying \(G\) to \(G'\)—as a series of graph-level operations that mirror the adjustments a user might perform on a scene. Although fundamental operations—adding objects, removing objects, and modifying relationships—are conceptually simple, each may result in conflicts and requires precise management, as illustrated in \autoref{fig:teaser}. We show that these operations can be reduced to inter-object relationship prediction in an autoregressive fashion.

\paragraph{Node Addition.} 
Adding a new object node into the scene graph involves more than a straightforward insertion. To ensure the added node follows the constraints in ~\eqref{eq:reasonable_sg}, predominantly the commonsense constraint \(P_1\), the \emph{Stitch} process requires accurate inter-object reasoning and the establishment of relationships between the new node and all existing nodes.
We denote a new node as \(v\). Define
\(
\bar{e} \subseteq \{\,\{v, v_i\} : v_i \in V\,\}
\)
as the set of edges between \(v\) and each \(v_i\).
Node addition is then formulated as constructing a new graph
\[
G' = \Bigl(V \cup \{v\},\, E \cup \bar{e}\Bigr) \in \mathbb{G}.
\]
Intuitively, every edge in $\bar{e}$ should not cause conflicts. This can be achieved by sequentially predicting inter-object relationships using an autoregressive approach.

\paragraph{Node Removal.}
Removing a node is a straightforward operation: removing the node and its associated edges, listed as a \emph{Cut} step. Let the node to be removed be \(v \in V\), and the associated edges are
\(
\bar{e} \subseteq \{\,\{v, v_i\} : v_i \in V \setminus \{v\}\,\}
\)
, where \(v_i\) are the connected nodes. Thus, it is formulated as constructing a new graph
\[
G' = \Bigl(V \setminus \{v\},\, E \setminus \bar{e}\Bigr) \in \mathbb{G}.
\]

\paragraph{Edge Change.}
Modifying spatial relationships is a challenging operation. This task requires case-by-case conflict detection and resolution to ensure compliance with the constraints in ~\eqref{eq:reasonable_sg}, particularly the spatial constraints \(P_2\). This often yields suboptimal results due to heavy engineering requirements and the lack of automatic commonsense reasoning. 
In contrast, we treat this problem as a \emph{Cut-And-Stitch} process:
\emph{Cut:} Temporarily isolate the node of interest by removing its incident edges.
\emph{Stitch:} Reinsert the node with its modified relationships and then infer and restore any missing edges, thereby reducing the problem to a node addition task.
Let the node of interest be \(v \in V\). Denote by \(\bar{e}' \subseteq E\) the set of edges originally incident to \(v\) and by 
\(
\bar{e} \subseteq \{\,\{v, v_i\} : v_i \in V \setminus \{v\}\,\}
\)
the set of predicted edges connecting \(v\) to the other nodes. The edge change operation is formulated as constructing a new graph
\[
G' = \Bigl( V,\, \left(E \setminus \bar{e}'\right) \cup \bar{e} \Bigr) \in \mathbb{G}.
\]

\bigskip
In summary, by decomposing each operation into a sequence of \emph{Cut} and \emph{Stitch} steps that can be solved by autoregressive inter-object relationship prediction, our formulation systematically addresses the challenges of scene graph manipulation while ensuring that structural coherence is rigorously maintained, i.e., \(G' \in \mathbb{G}\).

%% file: sec/5_method.tex
\section{SG-Tailor}

To address the challenges of node addition and edge change (illustrated in~\autoref{fig:pipeline}), we propose SG-Tailor—an autoregressive model that predicts inter-object relationships, $\bar{e}$, within a reasonable scene graph. By leveraging a sequential prediction strategy, SG-Tailor infers the appropriate relationships among nodes, ensuring that any newly added node is seamlessly integrated and that the overall graph remains coherent—that is, the manipulated graph $G' \in \mathbb{G}$.

\subsection{From Triplets to Tokens}

We formulate inter-object relationship reasoning as an autoregressive sequence generation task by encoding every entity and predicate in the dataset as a unique sequence of tokens. As illustrated in~\autoref{Problem}, given a reasonable scene graph $G \in \mathbb{G}$ with $|V|$ nodes and $|E|$ edges, we obtain $|E|$ triplets, each represented as $\{v_i, e_{i\ \to j}, v_j\}$. To enable autoregressive modeling based on a Transformer architecture, we first convert these triplets into tokens. Inspired by the quintuple token strategy introduced in~\cite{zhu2024mmt} for music processing, where each quintuple serves as a unified input to attention blocks, we adopt a similar structure. Specifically, we decompose each triplet into separate tokens representing class labels and instance IDs for the subject and object, along with the predicate, resulting in quintuple tokens:
\(
q_i = (v_i^{\text{cls}}, v_i^{\text{ind}}, v_j^{\text{cls}}, v_j^{\text{ind}}, e_{i\to j}).
\)
We then tokenize each quintuple $q_i$ into the token sequence $t_i$ and augment it with special tokens to explicitly guide the learning process:
\[
t_i = \Bigl( \mathbf{[BOQ]},\, {t}_{i}^1,\, {t}_{i}^2,\, {t}_{i}^3,\, {t}_{i}^4,\, \mathbf{[SEP]},\, {t}_{i}^p,\, \mathbf{[EOQ]} \Bigr),
\]
\noindent
where $t_i^1, t_i^2, t_i^3, t_i^4$ correspond to the tokens for $v_i^{\text{cls}}, v_i^{\text{ind}}, v_j^{\text{cls}}, v_j^{\text{ind}}$ respectively, and ${t}_{i}^p$ is the predicate token. These special tokens, namely $\mathbf{[BOQ]},\mathbf{[SEP]},\mathbf{[EOQ]}$, clearly delineate the boundaries between subject, object, and relation segments. After the token sequence $t_N$, we append a sequence-end token $\mathbf{[EOS]}$ (see~\autoref{fig:pipeline}.C).

This tokenization strategy preserves the categorical information, including entity classes and relation types, as well as instance information for each relationship. In other words, the semantic structure of the original scene graph is maintained within the linear token sequence. 
By encoding triplets as guided token sequences, we enable the model to attend to and learn from the rich structure of the scene graph sequentially.

\subsection{ Next-Token Learning}
At the core of SG-Tailor lies a decoder-only transformer $\phi$ that is trained to generate the scene graph token sequence one token at a time. During training, $\phi$ is supervised by every token inside of $t_i$ to learn in diverse domains, including both objects and their relationships. Hence, masking is applied at the individual token level rather than at the level of the entire triplet sequence $t_i$, as shown in~\autoref{fig:pipeline}.C. In Sec.~\ref{abla}, we demonstrate that it enhances performance by providing diverse supervision rather than solely focusing on the supervision of relationships.
\paragraph{Modeling}
Specifically, $\phi$ processes input sequences comprising tokens from previous $\{t_1, t_2, \dots, t_i\}$ and partial tokens $t_{i+1}^{0:c}, 0 \leq c < |t_{i+1}|$, where $|t_{i+1}|$ denotes the length of the token sequence, to predict the next token $t_{i+1}^{c+1}$. Formally, at each prediction step, we have:
\begin{equation}
P\left(t_{i+1}^{c+1}\middle|\,t_1, t_2,\dots, t_i, t_{i+1}^{0:c}\right),
\quad \forall i \leq N-1,\,
\end{equation}
This autoregressive modeling enables  $\phi$ to capture intricate dependencies at both the entity level and relationship level across the entire sequence, thus enhancing the reasoning capability on inter-object relationships.

\paragraph{Training Objective}
We adopt a categorical cross-entropy loss across the entire vocabulary to train the model:
\begin{equation}
\mathcal{L} = -\sum_{t=1}^{T}\log\frac{\exp\left(z_{t, y_t}\right)}{\sum_{w \in \mathcal{V}}\exp\left(z_{t, w}\right)},
\end{equation}
\noindent
where $T$ is the total number of tokens in the sequence, $\mathcal{V}$ represents the token vocabulary, $y_t$ denotes the ground-truth token at step $t$, and $z_{t,w}$ indicates the logit (unnormalized score) assigned by the model to token $w$ at step $t$. This loss guides the model to accurately predict each token, ensuring the effective autoregressive modeling of scene graph structures.

\subsection{ Next-Relationship Reasoning}
During inference, SG-Tailor performs autoregressive reasoning to predict the inter-object relationship between a query node and another node, conditioning each prediction on the previously established graph connections. Given contextual token sequences $T=\left\{t_i~|~i=1,...,N \right\}$ representing the existing graph, the model computes the conditional probability on the query tokens consisting of the query node and the other nodes to have 
\begin{equation}
    t_{i+1}^* = \Bigl( \mathbf{[BOQ]},\, {t}_{i}^1,\, {t}_{i}^2,\, {t}_{i}^3,\, {t}_{i}^4,\, \mathbf{[SEP]} \Bigr).
\end{equation}
Then, the reasoning can be formulated as
$
    P({{t}_{i}^p} \mid T,t_{i+1}^*).
$
This formulation enables the model to capture the intricate interdependencies among objects, ensuring that each new prediction respects the existing graph structure.

During every inference step, the subject and object tokens are queried as described in ~\autoref{fig:pipeline}.D, and the model predicts the next possible token using a top-\(p\) (nucleus) sampling strategy \cite{p-sampling}, in which the probability distribution is first sorted in descending order, and then only the smallest set of tokens whose cumulative probability exceeds a predefined threshold \(p = 0.7\) is considered for sampling. By the end of the inference, the scene graph $G' \in \mathbb{G}$ is built from the tokens.

%% file: sec/6_experiment.tex
\section{Experiments}
\subsection{Datasets}
We evaluate our method quantitatively and qualitatively
on four datasets, 3RScan~\cite{3rscan}, 3D-FRONT~\cite{3dfront}, SG-Bot~\cite{zhai2024sg}, and SceneVerse~\cite{sverse} with different motivations. Both 3RScan and 3D-FRONT are widely used as benchmarks for scene-graph-based 3D scene generation~\cite{dhamo2021graph, commonscenes}, which allows qualitative evaluation of our method on the 3D scene generation downstream task. Experiments on SceneVerse37K, a subset of the complete SceneVerse dataset, show the scalability of our method in large-scale scenarios.\\

Due to the absence of source-target 3D scene pairs in 3RScan and 3D-FRONT, we evaluate the user preference ranking of our method compared to a baseline based on SceneGraphNet~\cite{scenegraphnet}, the naive manipulation baseline described in~\cite{dhamo2021graph}, and the response from an instruction fine-tuned large language model Llama-3.3-70B-Instruct~\cite{grattafiori2024llama}. In addition to scene manipulation, we train our method on the SG-Bot dataset~\cite{zhai2024sg} to evaluate how our method facilitates scene-graph-based robotic manipulation. 

Training data is generated by encoding scene graph labels into sequences of tokens from our vocabulary.

\subsection{Implementation Details} \label{sec:implementation_details}
We use Llama \cite{llama} layers as the autoregressive transformer in our model, with a hidden size of 768 and 12 attention heads. We pad the scene graph sequence to the context length $1024$ and use a cosine learning rate scheduler with an initial learning rate of $5 \times 10^{-4}$ and a weight decay of $1 \times 10^{-2}$. The batch size is set to 16 in all our experiments, and we train our models for 50 epochs, employing early stopping. Unless otherwise specified, we predict all relationships using nucleus sampling \cite{p-sampling} with a p-value of 0.7. We augment our data by randomly shuffling each scene graph three times.

\subsection{Baselines}
\noindent\textbf{Naive Manipulation Baseline (Naive).}
Following~\cite{commonscenes,zhai2024echoscene}, we propose a baseline based on ~\cite{dhamo2021graph}. It simply modifies a target edge without any manipulation of the rest of the graph. Since this baseline cannot reason relationships, we remove it from node addition tasks.

\noindent\textbf{SceneGraphNet (SGNet).}
SceneGraphNet is a Message-Passing Neural Network inspired by GNN \cite{scenegraphnet}. We modify SGNet as a baseline. Since SGNet uses class labels, location, and dimension information to predict the likelihood over classes at a query location, we modify the encoder part to only process class information, keep the message-passing mechanism, and build two MLP decoders for our tasks of node addition and edge change, respectively.

\noindent\textbf{LLM (Llama-3.3).}
We use the latest Llama model, the Llama-3.3-70B-Instruct~\cite{grattafiori2024llama}, as our baseline. Specifically, we prompt the model with the task description, all the triplets describing a given scene, and the specific relationship we aim to modify. The LLM then processes this information and generates a revised scene graph based on our prompt. A detailed prompt template used for the edge change task on the dataset 3D-FRONT can be found in the supplementary. 

\subsection{Evaluation Metrics} 
We evaluate our method with two types of quantitative metrics: the ranking-based metrics and the scene graph consistency metrics.

\paragraph{Ranking-based Evaluation Metrics.} Since the accuracy of a predicted edge depends on scene context and multiple valid answers may exist, we adopt ranking-based metrics from the knowledge graph community \cite{TransE}. Please refer to the supplementary for the equations for these metrics.

\noindent\textbf{Mean Rank (MR)}
calculates the average ranking position of correct predictions, where lower ranks indicate better performance; outliers can skew results.

\noindent\textbf{Mean Reciprocal Rank (MRR)}
averages the reciprocal rank of the first correct prediction, emphasizing early correct answers and penalizing lower-ranked, delayed correct predictions.

\noindent\textbf{Hit@K}
computes the proportion of queries where the correct answer appears within the top K predictions, reflecting practical recommendation performance.

\begin{table*}[ht]
    \centering
    \resizebox{0.8\textwidth}{!}{
        \begin{tabular}{lllccccc}
            \toprule
            Method  & Dataset & MR $\downarrow$ & MRR $\uparrow$ & Hit@1 $\uparrow$ & Hit@3 $\uparrow$ & Hit@10 $\uparrow$\\ 
            \midrule
            SGNet~\cite{scenegraphnet} & \multirow{2}{*}{3RScan~\cite{3rscan}} & 3.987 & 0.572 & 0.398 & \textbf{0.683} & 0.941 \\
            Ours &  & \textbf{3.764} & \textbf{0.624} & \textbf{0.451} & 0.681 & \textbf{0.953} \\
            \midrule
            SGNet~\cite{scenegraphnet} & \multirow{2}{*}{SceneVerse37K~\cite{sverse}} & 5.922 & 0.367 & 0.199 & 0.426 & 0.882 \\
            Ours &  & \textbf{5.623} & \textbf{0.374} & \textbf{0.210} & \textbf{0.507} & \textbf{0.92} \\
            \midrule
            SGNet~\cite{scenegraphnet} & \multirow{4}{*}{3D-FRONT~\cite{3dfront}} & 4.103 & 0.486 & 0.271 & 0.593 & 0.943 \\
            Ours (GPT-2) &  & 4.113 & 0.472 & 0.273 & \textbf{0.604} & 0.947 \\
            Ours (Next-Rel) &  & 4.113 & 0.476 & 0.282 & 0.603 & 0.951 \\
            Ours &  & \textbf{3.613} & \textbf{0.498} & \textbf{0.305} & 0.591 & \textbf{0.983} \\
            \bottomrule
        \end{tabular}
    }
    \caption{Performance metrics across models and datasets. Best results are in bold}
    \label{tab:rank}
\end{table*}

\paragraph{Scene Graph Consistency Metric.}

Taking the intuition that the presence of a cycle signifies a spatial contradiction within the scene graph, we identify spatial conflicts through Algorithm 1 in supplementary material, a simple graph loop detection algorithm based on depth-first search. we convert left and behind triplets into their right and front counterparts and then detect cycles in the right and front relationships. 
This experiment is conducted among the naive approach, Llama-3.3-70B-Instruct, SceneGraphNet, and our approach by modifying spatial relations (left, right, front, behind) to generate new scene graphs.

\begin{table}[ht]
    \centering
    \begin{tabular}{lccc}
        \toprule
        Method & right cycle $\downarrow$ & front cycle$\downarrow$ & total$\downarrow$\\
        \midrule
        Naive &  19.19\% & 19.73\% & 38.92\%\\
        Llama-3.3~\cite{grattafiori2024llama} & 34.48\%&31.03\%&62.07\%\\
        SGNet~\cite{scenegraphnet} &  7.37\% & \textbf{0.0} & 7.37\%\\
        SG-Tailor (Ours) & \textbf{1.05\%} & \textbf{0.0} & \textbf{1.05\%}\\

        \bottomrule
    \end{tabular}
    \caption{\textbf{Cycle rate.} The percentage of scene graphs that have either a right or front cycle. The best results are in bold. } 
    \label{tab:cycle}
\end{table}

\begin{table}[ht]
    \centering
    \begin{tabular}{lcc}
        \toprule
        Method & manipulation$\uparrow$ & addition$\uparrow$ \\
        \midrule
        Naive &  1.72\% & - \\
        Llama-3.3~\cite{grattafiori2024llama} &  9.48\% & 18.97\% \\
        SGNet~\cite{scenegraphnet} &  18.10\% & 33.33\% \\
        SG-Tailor (Ours) & \textbf{70.69\%} & \textbf{47.70\%} \\

        \bottomrule
    \end{tabular}
    \caption{\textbf{Top-1 rate.} The percentage of participants who consider each method across different tasks. The best results are in bold.}
    \label{tab:user_study}
\end{table}

\begin{figure*}[ht] 
    \centering          
    \includegraphics[width=0.98\textwidth]{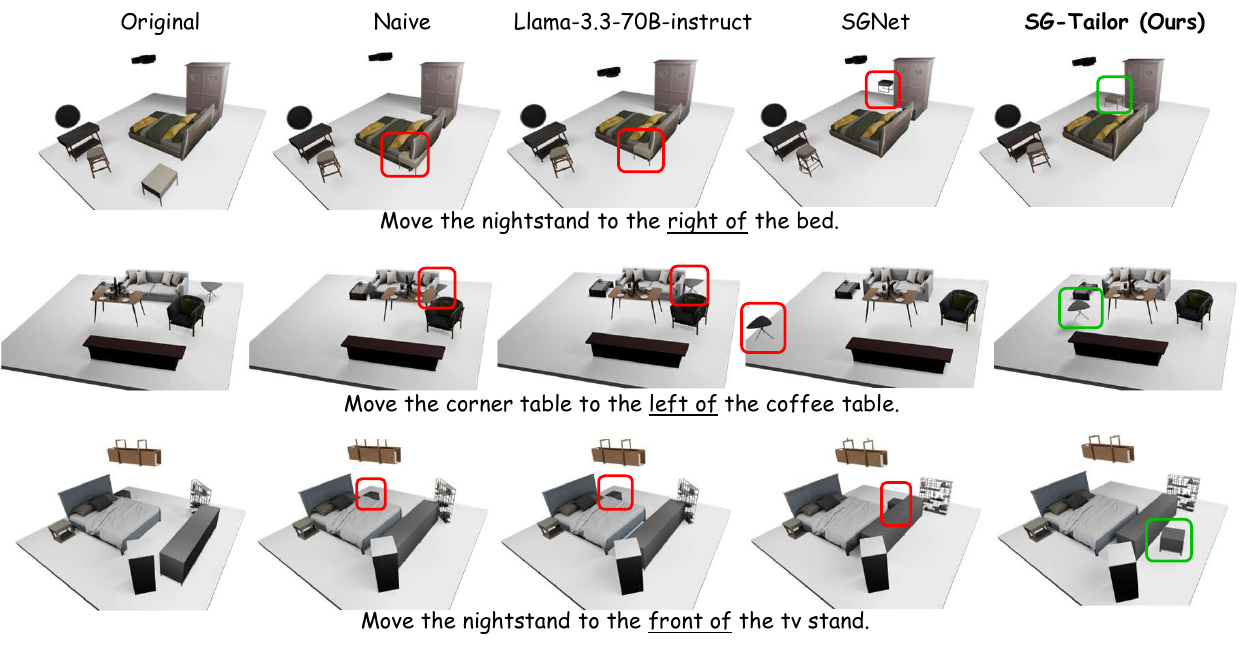} 
    \caption{\textbf{Qualitative comparison.} We assess the quality of scene graph manipulation across different methods by generating the corresponding scenes using the Graph-to-3D model.~\cite{dhamo2021graph}.} 
    \label{fig:comparison} 
\end{figure*}

\begin{figure*}[ht] 
    \centering          
    \includegraphics[width=0.98\textwidth]{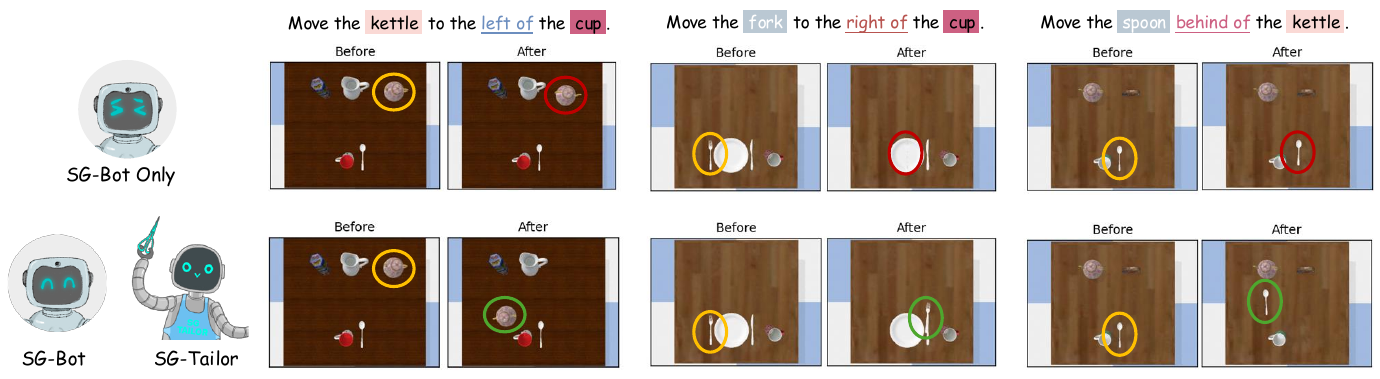} 
    \caption{\textbf{Qualitative comparison of SG-Bot w/ and w/o SG-Tailor.} We show three examples of SG-Tailor facilitating the robotic manipulation tasks. More examples can be found in the supplementary.} 
    \label{fig:sgbot_comparison} 
\end{figure*}

\subsection{Ablation Study} 
\label{abla}
We also evaluate our method against two other variants based on the same dataset and evaluation metrics in Table~\ref{tab:rank}.  Specifically, we test the following variants: (1) \textbf{SG-Tailor (GPT-2).} We train the GPT-2 layers~\cite{gpt2} of the same depth and width on the same next token prediction task. (2) \textbf{SG-Tailor (Next-Rel)} We train SG-tailor with loss label masks on all subjects and objects, effectively guiding the model to only learn to predict the predicates. We found that our method outperforms all these variants.

\subsection{Quantitative Results}
We compare the performance of our method to the baseline method on several different datasets.

\noindent \textbf{Performance on 3RScan:} On the 3RScan dataset, our method achieves notable improvements over SGNet. Specifically, our approach reduces the MR from 3.9872 to 3.7636, indicating a more accurate ranking of true relationships. Furthermore, our method boosts the MRR from 0.5719 to 0.6241 and the Hit@1 score from 0.3974 to 0.4511. These enhancements suggest that our method is significantly better at prioritizing correct relationships in the ranking process.

\noindent \textbf{Performance on SceneVerse:} Our full SG-Tailor model performs better than the baseline on this large-scale and challenging dataset on all metrics with a \textbf{5.52\%} relative increase on Hit@1. We argue that SGNet is based on GRU and has less generalization ability.

\noindent \textbf{Performance on 3D-FRONT:} On the 3D-FRONT dataset, our final configuration delivers an \textbf{11.9\%} reduction in MR (from 4.1029 to 3.6129) and achieves superior Hit@1 (0.3048 vs. 0.2708) and Hit@10 (0.9833 vs. 0.9433) scores compared to SGNet. While the overall MRR and Hit@3 scores remain competitive, our GPT-2 variant further excels on Hit@3 (0.6121), highlighting the potential benefits of incorporating language model features for specific ranking aspects.

\noindent \textbf{Scene Graph Consistency Result:} In addition to ranking metrics, we evaluate the cycle rate of the manipulated scene graphs, as shown in Table~\ref{tab:cycle}. Notably, over 30\% of the scene graphs generated by the naive approach contain significant contradictions, whereas our method effectively mitigates these inconsistencies, achieving a remarkably low cycle rate of just 1\%.
We infer that the exceptionally high cycle rate observed for the LLM is likely due to the fact that instruction-tuned LLMs are prone to noise\cite{llm_noise}.
\noindent Overall, these quantitative results clearly demonstrate that conditioning on the entire scene graph context not only improves the ranking accuracy of relationship predictions but also yields more coherent and spatially consistent scene graphs. The consistent performance gains across different datasets and metrics underscore the robustness and effectiveness of our approach.

%% file: sec/7_downstream_task.tex
\section{Downstream Applications}
\subsection{Scene Manipulation}

In this section, we evaluate our method on the downstream task of scene manipulation. Operating entirely at the scene graph level, our model integrates seamlessly with existing scene-graph-based 3D scene generation frameworks \cite{dhamo2021graph, commonscenes, zhai2024echoscene}. We train SG-Tailor on the 3D-FRONT dataset — the same dataset used for the downstream module Graph-to-3D—and perform scene graph manipulation before feeding the graph into Graph-to-3D for 3D rendering. We compare the rendered results of our method with those of SGNet, LLM, and a naive edge-change approach implemented in Graph-to-3D. Additionally, we conduct a perceptual study to gauge user preferences: the study evaluates four methods on the edge-change task and, for the node addition task, excludes the naive method, which lacks the spatial inference ability to integrate a new node into the scene.

\paragraph{Qualitative Results.} Our scene graph manipulation method demonstrates strong potential for enhancing 3D scene generation. By modifying the relationships within a scene graph, our approach can generate a diverse range of 3D scenes. To assess the quality of the manipulated graphs, we render the corresponding 3D scenes using Graph-to-3D (see Figure \ref{fig:comparison}). Our method effectively captures the global context, resulting in scene graphs that are both accurate and coherent. In contrast, alternative methods struggle to enforce object constraints and predict new relationships accurately, often leading to inconsistent object placements and unrealistic arrangements in the scenes.

\paragraph{User Study.} To compensate for the lack of ground-truth 3D scenes after manipulation, we conducted a perceptual evaluation study with 30 randomly selected participants. In the edge change part, participants are presented with the original scene and scenes generated by four different methods: naive, SGNet, LLM, and SG-Tailor. In the node addition part only results from  LLM, SGNet, and SG-Tailor are compared. Participants rank the scenes according to how well the changes in the scenes reflect the perceptual similarity to the provided task description. The summarized results in Table \ref{tab:user_study} demonstrate that our method outperforms all others in both node addition and edge change tasks.

\subsection{Robotic Manipulation}
Beyond scene manipulation, we integrate SG-Tailor into robotic manipulation tasks within tabletop environments. Our approach builds upon SG-Bot~\cite{zhai2024sg}, a robotic manipulation method that leverages graph generative models to determine goal states. While SG-Bot excels at generating precise target configurations, it often encounters difficulties rearranging objects when conflicts emerge in the scene graph following relationship updates. As illustrated in \autoref{fig:sgbot_comparison}, SG-Tailor overcomes these challenges by employing our novel \emph{Cut-And-Stitch} strategy combined with robust inter-object reasoning. This integration produces more coherent and context-aware scene representations, significantly enhancing SG-Bot’s planning accuracy and execution efficiency. Additional qualitative results, including a detailed image stream, are presented in the Supplementary.

%% file: sec/8_conclusion.tex
\section{Limitation}
We observed that our SG-Tailor model does not perform well when the training scene graphs are overly noisy. For example, an excessive number of non-spatial relationships, such as \textit{same material as}, can confuse the model, leading it to prioritize predicting these relationships rather than focusing on spatial constraints.  
Another potential area for improvement lies in our current approach to manipulating objects within the scene graph. When modifying an object, we remove all edges connected to its node and predict new edges. However, this approach may be simplistic and brute-force, as it disregards contextual information from the original connections, potentially leading to undesired scene graph modifications.

\section{Conclusion}
We present SG-Tailor, an autoregressive model for scene graph manipulation that addresses the limitations of existing methods by considering the global scene context and commonsense relationships. Unlike prior approaches that modify individual edges in isolation, our model predicts new relationships conditioned on the entire scene graph, ensuring spatial consistency. Experimental results demonstrate that SG-Tailor generates more realistic and contradiction-free 3D scenes. This work opens new directions for intelligent and context-aware scene generation, with potential applications in areas such as diverse 3D scene generation and robotics manipulation.

%% file: sec/supplementary.tex
\clearpage
\section*{Supplementary Materials}

\section{Metrics}
As discussed in the main paper, we evaluate our method with ranking-based metrics: mean rank, mean reciprocal rank, and Hits@k. The detailed definition is as follows:

\noindent\textbf{Mean Rank (MR)}
\begin{equation}
    \label{eq:mr}
    \mathbf{MR} = \frac{1}{N}\sum_{i=1}^{N} \text{rank}_i.
\end{equation}

\noindent\textbf{Mean Reciprocal Rank (MRR)}
\begin{equation}\label{eq:mrr}
\mathbf{MRR} = \frac{1}{N}\sum_{i=1}^{N} \frac{1}{\text{rank}_i}.
\end{equation}

\noindent\textbf{Hit@k}
\begin{equation}\label{eq:hit@k}
\mathbf{Hit@k} = \frac{1}{N}\sum_{i=1}^{N} \mathbb{1}\{\mathrm{rank}_i \le k\}.
\end{equation}

\section{Cycle Detect Algorithm}
We identify spatial conflicts through Algorithm~\ref{alg:cycle_detect}, which is a graph loop detection algorithm based on depth-first search (DFS). Specifically, we convert left and behind triplets into their right and front counterparts and then detect cycles in the right and front relationships. 


\begin{algorithm}[H]
\caption{DFS-based Cycle Detection in a Directed Graph}
\label{alg:cycle_detect}
\begin{algorithmic}[1]
\Procedure{DetectCycle}{$G$}
    \State $visited \gets \emptyset$
    \State $recStack \gets \emptyset$
    \For{each vertex $v$ in $G$}
        \If{$v \notin visited$}
            \If{\Call{DFS}{$v$, $visited$, $recStack$, $G$}} 
                \State \Return \textbf{true} \Comment{Cycle detected}
            \EndIf
        \EndIf
    \EndFor
    \State \Return \textbf{false} \Comment{No cycle found}
\EndProcedure
\end{algorithmic}
\end{algorithm}

\section{LLM Manipulation Prompt}\label{prompt}
We provide the prompts that we use for the edge change task with Llama-3.3-70B-Instruct~\cite{grattafiori2024llama}.

\lstset{
    basicstyle=\ttfamily\small,
    numbers=none,
    frame=single,
    breaklines=true,
    tabsize=2,
    captionpos=b,
    showstringspaces=false
}
\begin{lstlisting}
[Context]
You are a helpful assistant whose task is 
to manipulate a node in a scene graph. 
The scene graph is represented as 
triplets in the following format:
subject object relationship

Possible relationships are:
left
right
front
behind
standing_on
bigger_than
smaller_than

Input Format:
First line: The triplet of the 
node to be set.

Subsequent lines: Existing scene 
graph triplets (one per line).

Instructions:
Output Requirements:
Respond only with the complete scene 
graph triplets after adding new tripets.

Do not include any explanations or 
commentary in your output.

Removing Triplets:
Skip the first triplet, remove any 
triplets that contain the subject 
of the first triplet.

Adding Triplets:
Use the subject node of the first triplet 
as subject, and select at most 
4 other nodes as objects, 
and one of the possible relationships, 
form at most one triplet for each of the 
selected object, and add to the list. 

Only add triplets that are consistent 
with the spatial constraints of 
the existing scene graph.
Do not add triplets that are 
already present in the scene graph.
Do not add triplets that are contradictory 
to the existing scene graph.
Do not add triplets that are redundant.

Spatial Relationships:
Ensure that the updated scene graph 
has no contradictions in 
spatial relationships. 
(A contradiction is defined as two or more 
triplets that imply mutually exclusive 
spatial configurations.)

Output Format:
Your response should include the 
entire updated scene graph, 
in the exact order specified 
by the input plus any new valid triplets.
Do not include any extra text, formatting, 
or explanations.
Respond one triplet per line.

Example:
Example Input:
chair_1 desk_1 right
chair_1 floor_1 standing_on
desk_1 floor_1 standing_on
chair_1 desk_1 left
chair_2 desk_1 right
chair_2 floor_1 standing_on

Example Output:
desk_1 floor_1 standing_on
chair_2 floor_1 standing_on
chair_1 desk_1 right
chair_1 floor_1 standing_on
chair_1 chair_2 left
[/Context]
\end{lstlisting}

\section{Statistics of user study}
We show that across two tasks, there is a clear trend favoring our method. Our method, combined with the downstream 3D scene generation module, provides a solid framework for 3D scene manipulation.

\begin{figure}[htbp]
    \centering
    \begin{tikzpicture}
        \begin{axis}[
            ybar stacked,
            bar width=20pt,
            symbolic x coords={Naive, Llama, SGNet, SG-Tailor},
            xtick=data,
            ymin=0, ymax=100,
            ylabel={Percentage(\%)},
            legend style={at={(0.5,-0.15)}, anchor=north, legend columns=-1}
        ]
            \addplot coordinates {(Naive,1.72) (Llama,9.48) (SGNet,18.1) (SG-Tailor,70.69)};
            \addplot coordinates {(Naive,8.62) (Llama,12.93) (SGNet,57.76) (SG-Tailor,20.69)};
            \addplot coordinates {(Naive,33.62) (Llama,53.88) (SGNet,8.19) (SG-Tailor,4.31)};
            \addplot coordinates {(Naive,56.03) (Llama,23.7) (SGNet,15.95) (SG-Tailor,4.31)};
            \legend{Rank 1, Rank 2, Rank 3, Rank 4}
        \end{axis}
    \end{tikzpicture}
    \caption{Statistics of the user study in the manipulation task}
    \label{fig:user_study_mani} 
\end{figure}
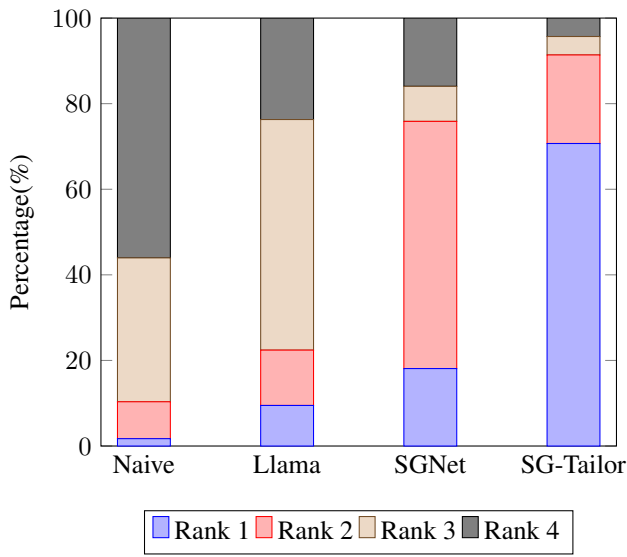

\begin{figure}[htbp]
    \centering
    \begin{tikzpicture}
        \begin{axis}[
            ybar stacked,
            bar width=20pt,
            symbolic x coords={Naive, Llama, SGNet, SG-Tailor},
            xtick=data,
            ymin=0, ymax=100,
            ylabel={Percentage(\%)},
            legend style={at={(0.5,-0.15)}, anchor=north, legend columns=-1}
        ]
            \addplot coordinates {(Llama,18.96) (SGNet,33.33) (SG-Tailor,47.7)};
            \addplot coordinates {(Llama,29.89) (SGNet,30.46) (SG-Tailor,39.65)};
            \addplot coordinates {(Llama,51.15) (SGNet,36.2) (SG-Tailor,12.64)};
            \legend{Rank 1, Rank 2, Rank 3}
        \end{axis}
    \end{tikzpicture}
    \caption{Statistics of the user study in the addition task}
    \label{fig:user_study_add} 
\end{figure}
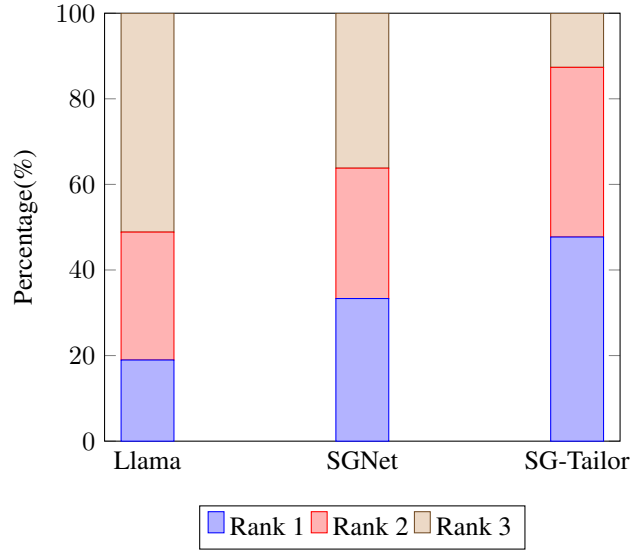

\begin{figure*}[t!] 
    \centering          
    \includegraphics[width=0.98\textwidth]{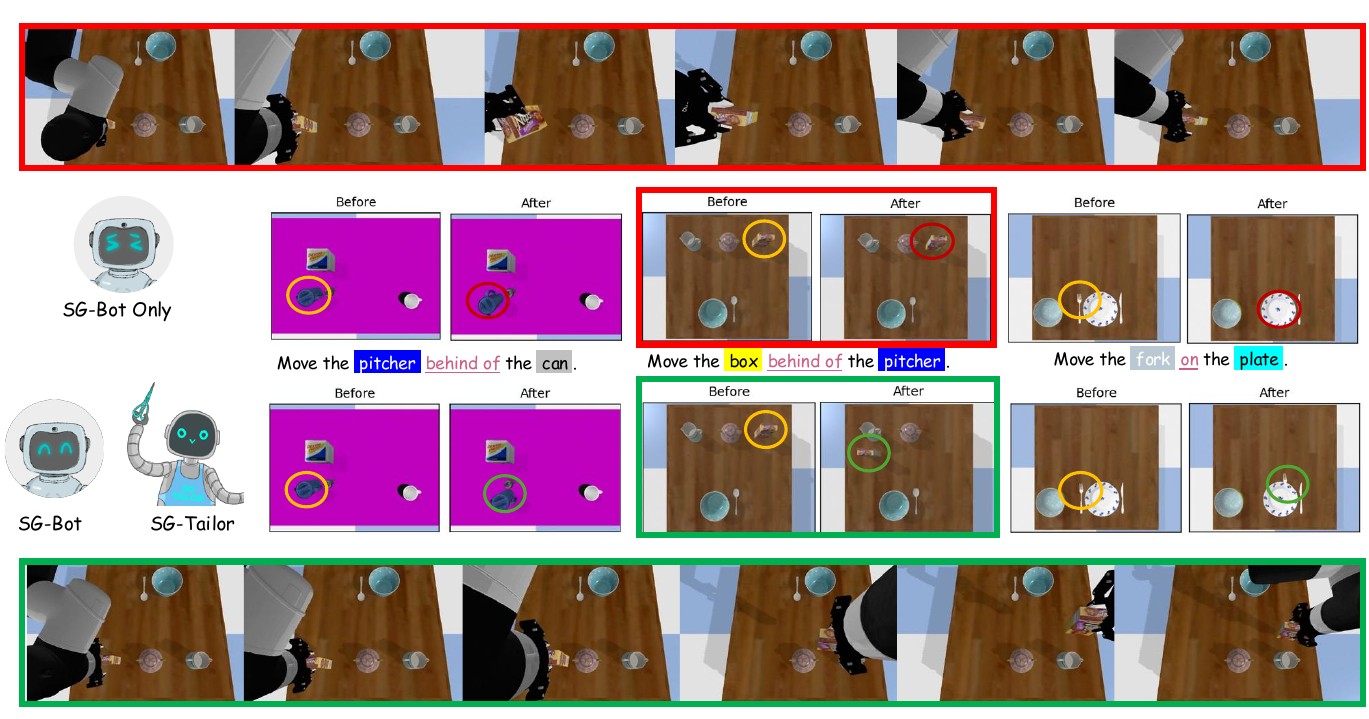} 
    \caption{More qualitative comparison of SG-Bot w/ and w/o SG-Tailor.} 
    \label{fig:sgbot_supp} 
\end{figure*}

\section{Additional Robotic Manipulation Results}\label{sgbot_supp}
We show additional performance when combining SG-Bot with Sg-Tailor in~\autoref{fig:sgbot_supp}, further showcasing the conflict-resolving ability of SG-Tailor. The above red stream in~\autoref{fig:sgbot_supp} shows the original SG-Bot is able to pick up the box, but due to the conflicts in the scene graph, the generation model fails, so the target location is still around the starting pose. In contrast, SG-Tailor can help by resolving the conflicts in the scene graph, so the generative model works again, thereby able to rearrange objects (see green stream).

\section{User Study Interface}
\begin{figure*}[ht] 
    \centering          
    \includegraphics[width=0.3\textwidth]{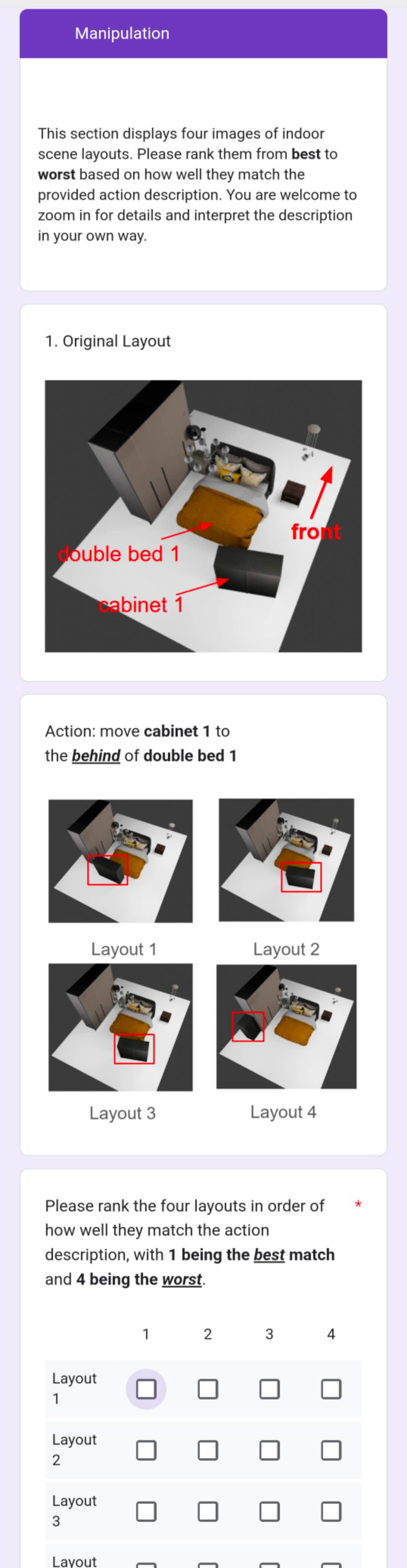} 
    \caption{\textbf{User interface for the perceptual user study.}} 
    \label{fig:user_interface} 
\end{figure*}

In this section, we detail the user study conducted to evaluate participants' perceptions of the rendered 3D scene image rankings. The study was structured into two parts, focusing on the node addition and edge change tasks. In each part, participants provided conceptual evaluations of the rankings based on their personal interpretations. To ensure unbiased responses, the images were presented in a completely randomized order with only minimal instructions provided, as shown in \autoref{fig:user_interface}.

